%% file: main.tex
\documentclass[journal,twoside,web]{ieeecolor}
\usepackage{jsen} 
\usepackage{amsmath,amssymb,amsfonts}
\usepackage{algorithmic}
\usepackage{graphicx} 
\usepackage{wrapfig}
\usepackage{cite}   
\usepackage{textcomp}
\usepackage{framed}
\usepackage{float} 
\usepackage{cuted}
\usepackage{xcolor}
\usepackage{placeins} 
\usepackage[utf8]{inputenc}
\usepackage{pgfplots}
\DeclareUnicodeCharacter{2212}{−}
\usepgfplotslibrary{groupplots,dateplot}
\usetikzlibrary{patterns,shapes.arrows}
\pgfplotsset{compat=newest}
\usepackage[font=small,labelfont=bf]{caption} 
\usepackage{mwe} 
\usepackage{adjustbox} 
\usepackage{lipsum}  
\usepackage{tikz}
\usepackage{multirow}   
\usepackage{hyperref}  
\usepackage{pifont} 
\usepackage{subcaption} 
\usepackage{booktabs}
\usepackage{tabularx}
\usepackage{siunitx}  
\usepackage{colortbl}

\usepackage[markup=underlined, commentmarkup=margin, final]{changes}
\definecolor{changed}{RGB}{0, 102, 204}     
\definecolor{deleted}{RGB}{204, 0, 0}       
\definecolor{comment}{RGB}{0, 153, 0}       
\definecolor{forestGreen}{RGB}{34, 139, 34}
\definecolor{firebrick}{RGB}{178, 34, 34}
\newcommand{\myxmark}{\color{firebrick}{×}}
\newcommand{\mycheckmark}{\color{forestGreen}{\checkmark}}

\def\BibTeX{{\rm B\kern-.05em{\sc i\kern-.025em b}\kern-.08em
    T\kern-.1667em\lower.7ex\hbox{E}\kern-.125emX}}
\definecolor{abstractbg}{rgb}{0.89804,0.94510,0.83137}
\begin{document}
\title{PicoSAM3: Real-Time In-Sensor Region-of-Interest Segmentation}
\author{
Pietro Bonazzi$^{\star\dagger}$,
Nicola Farronato$^{\star\dagger\ddagger}$,
Stefan Zihlmann$^{\dagger}$,
Haotong Qin$^{\dagger}$,
and Michele Magno$^{\dagger}$%
\thanks{$^{\star}$ Equal contribution}%
\thanks{$^{\dagger}$ ETH Z\"urich, Z\"urich, Switzerland.}%
\thanks{$^{\ddagger}$ IBM Research, Z\"urich, Switzerland.}%
\thanks{This work was submitted for review on the 30.01.2026 and was accepted on the 04.06.2026. It was funded by the Swiss National Science Foundation under Grant 219943 and by the European Union’s HORIZON-MSCA-2022-DN-01 under Grant 101119554. For correspondences please contact: pbonazzi@ethz.ch.}
}

\IEEEtitleabstractindextext{%
\fcolorbox{abstractbg}{abstractbg}{%
\begin{minipage}{\textwidth}%
\begin{wrapfigure}[18]{r}{4.05in}%
\includegraphics[width=4in]{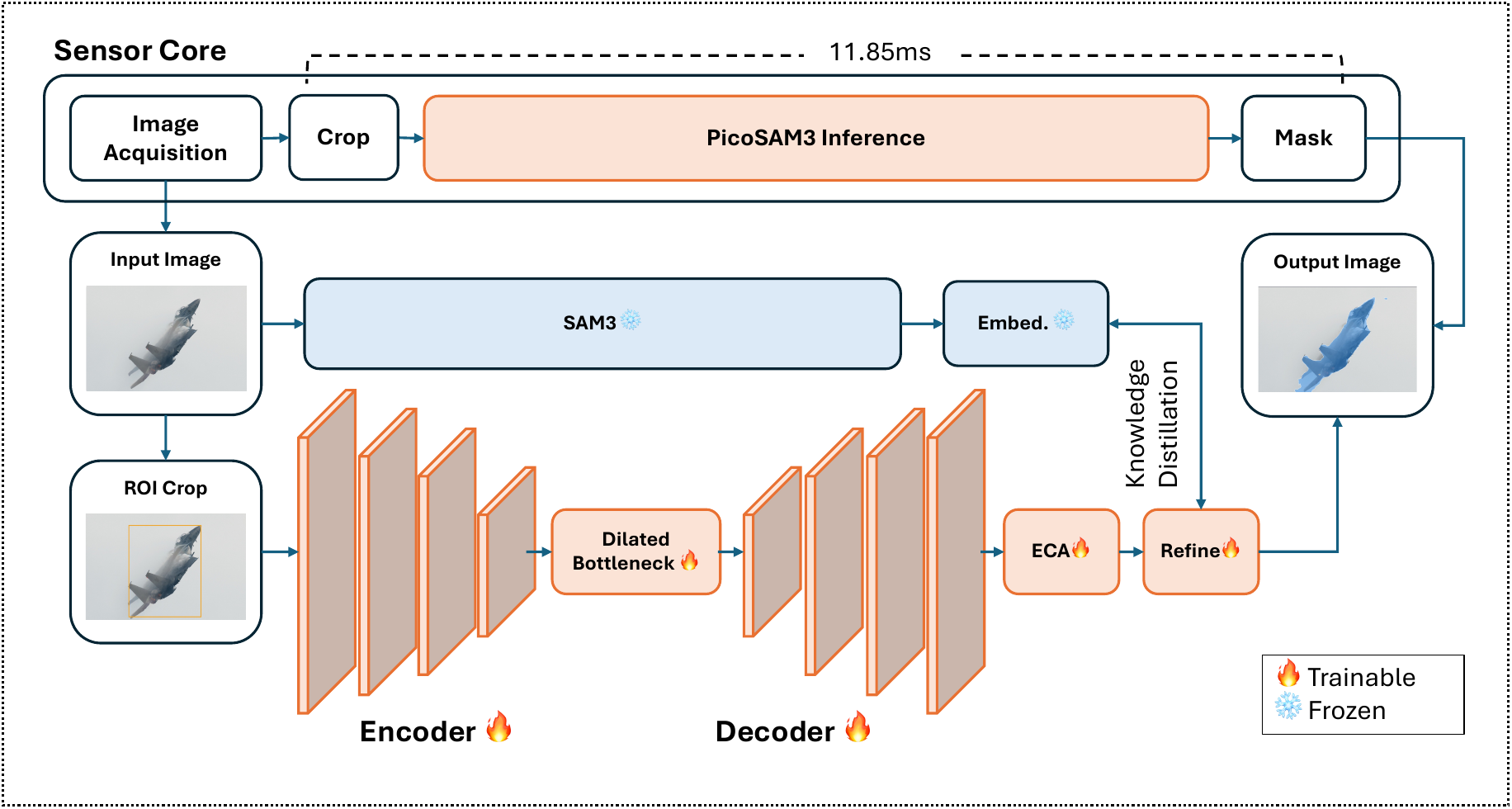}%
\label{fig:visual_abstract}
\end{wrapfigure}%


\begin{abstract} 
Real-time, on-device segmentation is critical for latency-sensitive and privacy-aware applications such as smart glasses and Internet-of-Things devices. We introduce PicoSAM3, a lightweight promptable visual segmentation model optimized for edge and in-sensor execution, including deployment on the Sony IMX500 vision sensor. PicoSAM3 has  1.3\,M parameters and combines a dense CNN architecture with region of interest prompt encoding, Efficient Channel Attention, and knowledge distillation from SAM2 and SAM3. On COCO and LVIS, PicoSAM3 achieves 65.45\% and 64.01\% mIoU, respectively, outperforming existing SAM-based and edge-oriented baselines at similar or lower complexity. The INT8 quantized model preserves accuracy with negligible degradation while enabling real-time in-sensor inference at 11.82\,ms latency on the IMX500, fully complying with its memory and operator constraints. Ablation studies show that distillation from large SAM models yields up to +14.5\% mIoU improvement over supervised training and demonstrate that high-quality, spatially flexible promptable segmentation is feasible directly at the sensor level.

\end{abstract}

\begin{IEEEkeywords}
Edge AI, In-sensor computing, Knowledge distillation, Low-latency vision, Promptable segmentation, Quantization, Real-time segmentation, Segment Anything Model, Smart sensors
\end{IEEEkeywords}
\end{minipage}}}

\maketitle

\section{Introduction}
\label{sec:introduction}
Recent advances in task-agnostic segmentation, spearheaded by Meta's Segment Anything Model (SAM) \cite{kirillov2023segment} and its successors SAM 2 \cite{ravi2024sam2} and SAM 3 \cite{carion2025sam3segmentconcepts}, have established a new standard for prompt-based visual understanding. This evolution has notably addressed the latency limitations of the original architecture; SAM 2, in particular, adopts a streamlined hierarchical backbone to extend segmentation capabilities to the video domain with improved throughput. Yet, despite these algorithmic refinements, the fundamental memory footprint and computational density of the transformer architecture remain prohibitive for extreme edge-sensing paradigms \cite{Fan_EfficientRemoteSensing, picosam2}. This presents a critical barrier for intelligent sensor nodes and low-power vision systems, such as battery-operated smart glasses \cite{Campanella_WearableSensor2024}, where the quadratic complexity of self-attention mechanisms continues to impede real-time execution.

Moreover, for resource-constrained scenarios, moving beyond centralized cloud processing toward near-sensor intelligence requires a radical reduction in model parameters to maintain operational longevity without compromising inference performance~\cite{Vitolo_Low-PowerDetection2022}.

To mitigate the communication bottlenecks inherent to cloud-centric AI, edge computing paradigms have emerged as a vital solution, enabling low-latency, privacy-preserving inference closer to the data source. By performing vision tasks locally, these systems circumvent the bandwidth constraints and jitter associated with offloading raw data \cite{moosmann2023ultraefficient, wang2020fann, giordano2022survey}. However, recent progress in semiconductor manufacturing is pushing this trend even further, shifting the locus of computation from external microcontrollers directly into the sensing node itself. This paradigm, known as in-sensor computing, allows for real-time perception immediately at the point of capture, eliminating data transmission overhead entirely.

This evolution is not limited to optical modalities; it represents a fundamental shift in sensor architecture across the spectrum. Driven by the synergy between multimodal smart sensors and innovative TinyML techniques, recent research has successfully embedded intelligence into inertial \cite{Kuhne_Resource-ConstrainedUAVs2025,Leiva_IMU2025}, olfactory \cite{Wei_GasConcentration2024,Cho_gas2024}, acoustic \cite{Verma_AviEar2024}, tactile \cite{zhou2024simulation}, and biomedical \cite{xie2024design} systems. Pushing this integration limit further, a new generation of devices now features embedded AI cores or dedicated hardware accelerators directly within the discrete sensor package. For instance, STMicroelectronics has introduced a Machine Learning Core (MLC) capable of executing decision trees directly within the MEMS sensor circuitry, allowing for a drastic reduction in the system's overall power budget \cite{MessinaQVAR_2025}. In the visual domain, a representative example is Sony's IMX500 \cite{imx500_sony_sensor, eki2021sonyIMX500, bonazzi2023tinytracker}, which integrates a CMOS image sensor vertically stacked with a dedicated edge AI processor \cite{zhou2020near}.

Despite these hardware advancements, in-sensor processing imposes constraints far more stringent than typical mobile or edge scenarios. For example, the IMX500 operates within a strictly bounded on-chip SRAM of less than 8 MB and supports a limited set of quantization-friendly operators. These physical limitations prevent existing promptable segmentation models such as TinySAM \cite{shu2023tinysam}, EdgeSAM \cite{zhou2023edgesam}, MobileSAMv2 \cite{zhang2023mobilesamv2}, and LiteSAM \cite{fu2024litesam} from being deployed. Such models typically exceed memory budgets, rely on unsupported operators (e.g., complex non-linear activations), or depend on the large-scale memory access patterns of transformer-based architectures. Furthermore, the lack of extensive cache hierarchies in these sensor-integrated AI cores makes the random memory access inherent to Self-Attention mechanisms a primary bottleneck for real-time throughput. Consequently, realizing promptable segmentation on such devices necessitates a shift from standard high-precision floating-point operations to ultra-low-bitwidth quantization~\cite{menghani2023efficient} and hardware-aware architectural design~\cite{qin2024mobilenetv4}.

Building on prior in-sensor segmentation work \cite{picosam2}, this paper introduces PicoSAM3, where a region of interest (ROI) prompted segmentation model explicitly designed for ultra-low-latency, in-sensor deployment. Unlike earlier approaches that relied on restrictive centered-point prompts, PicoSAM3 implements a spatially flexible ROI supervision strategy. This design encodes prompt information implicitly during training, thereby eliminating the computational overhead of inference-time prompt processing and enabling seamless integration with the hardware ROI capabilities of intelligent sensors like the Sony IMX500. To maximize representational power within strict edge constraints, PicoSAM3 replaces heavy transformer blocks with a dense CNN architecture augmented by a dilated bottleneck and lightweight Efficient Channel Attention (ECA) \cite{wang2020eca}. This approach ensures full compatibility with the limited operator set and 8MB memory budget of the IMX500 while maintaining high feature selectivity. Furthermore, we leverage a robust knowledge distillation pipeline transferring capabilities from the foundational SAM2 and SAM3 models to our compact student network.

The main contributions and results of this work are summarized as follows:
\begin{itemize}
\item We propose a hardware-aware architecture that operates in the 1.3-1.4 M parameter regime, employing INT8 quantization to achieve a 4x size reduction (to 1.31 MB) with negligible accuracy degradation.
\item Through novel distillation from SAM3, PicoSAM3 achieves state-of-the-art performance among edge-oriented models, reaching 65.45\% mIoU on COCO and 64.01\% on LVIS, surpassing existing baselines at comparable complexity.
\item We demonstrate real-time on-device execution on the Sony IMX500 vision sensor, achieving an end-to-end inference latency of 11.82 ms, confirming the feasibility of high-quality promptable segmentation at the extreme edge without cloud offloading.
\end{itemize}

\section{Related Work}
\label{sec:related_work}

The introduction of large foundation models such as SAM \cite{kirillov2023segment} has fundamentally altered the landscape of promptable visual segmentation. As detailed in the Zhang et al. survey \cite{zhang2023survey}, these models utilize massive transformer-based encoders to achieve zero-shot generalization across diverse domains, ranging from medical imaging to remote sensing. However, the substantial computational cost and memory footprint of the ViT~\cite{dosovitskiy2020image} backbone prohibit direct deployment on resource-constrained edge devices. Consequently, recent literature has prioritized methods to decouple segmentation capability from model size.

A primary research direction addresses architectural efficiency by physically replacing heavy backbones with lightweight alternatives. Approaches such as FastSAM \cite{zhao2023fast} and LiteSAM \cite{fu2024litesam} employ compact convolutional or hybrid encoders, significantly reducing parameter count. Complementary to these structural changes, knowledge distillation methods \cite{zhang2023faster, zhou2023edgesam, shu2023tinysam, picosam2} transfer the representational power of the large pretrained teacher to a compact student network, often bridging the accuracy gap through extensive dataset distillation. Distinct from these training-based approaches, training-free acceleration techniques \cite{liang2022expediting} exploit the spatial redundancy inherent in visual data. By dynamically merging or pruning less informative tokens during inference, these methods reduce the effective sequence length processed by the transformer without requiring parameter updates.

Finally, representational efficiency is addressed through model pruning \cite{chen20230} and post-training quantization (PTQ) \cite{lv2024ptq4sam, liu2023pqsam}, which reduce the precision of weights and activations to lower memory bandwidth usage. While effective in theoretical compression, these approaches often overlook the hardware-specific constraints of in-sensor computing, such as the lack of support for non-linear operations (e.g., Softmax, LayerNorm) and the requirement for strict integer-only arithmetic. This work addresses that gap by proposing a quantization-friendly architecture explicitly tailored for such extreme edge environments.

While the performance gap between CNN and transformer architectures is well documented at high resolution, the IMX500 hardware imposes a qualitatively different constraint. The Sony MCT target-platform-capability (TPC) op-set does not include integer implementations of Softmax (required by self-attention), LayerNorm (required in every transformer sub-layer), or GELU (used in transformer FFNs). Any model containing these operations fails MCT compilation entirely and cannot be deployed on the sensor. We empirically confirm this with MobileViT-XXS-Seg, a hybrid model equipped with a MobileViT-XXS encoder and the same PicoSAM3-style decoder trained with identical SAM3 distillation: MCT raises a quantisation error at the PTQ stage due to unsupported operations, while PicoSAM3 compiles and exports successfully. Despite the hybrid model having 5.7\,M parameters (vs.\ 1.37\,M), PicoSAM3 achieves higher mIoU (65.4\,\% vs.\ 63.1\,\%) at 96$\times$96 resolution. This demonstrates that within the set of architectures deployable on the IMX500, the purely convolutional design is not a compromise but the optimal choice.


Recent advances in PTQ attempt to compress SAM for edge deployment, yet they often lack validation on constrained hardware. For instance, PTQ4SAM \cite{lv2024ptq4sam} addresses the severe activation outliers in SAM's transformer blocks by employing a dual-stage quantization strategy that specifically isolates significant features to reduce quantization error. Similarly, PQ-SAM \cite{liu2023pqsam} utilizes Hessian-guided metrics to allocate mixed precision to sensitive layers, mitigating the impact of non-Gaussian distributions inherent to the ViT architecture. However, these methods primarily focus on theoretical compression rates rather than actual latency on microcontrollers, rendering them unsuitable for in-sensor development where memory and compute are strictly bounded. In contrast, Picosam3 strategically adopts a fully convolutional architecture to eliminate the activation outliers endemic to transformer-based models. This design choice ensures feature distributions remain compact and Gaussian-like, enabling near-lossless uniform quantization on low-power DSPs without the need for complex outlier suppression algorithms.

TinySAM \cite{shu2023tinysam} represents a significant advancement in distillation-based compression, reducing the computational overhead of SAM while maintaining zero-shot capabilities. The method employs a full-stage knowledge distillation framework that transfers feature information from the heavy ViT teacher to a lightweight student at multiple network depths. To further enhance robustness, TinySAM incorporates an online hard prompt sampling strategy. Unlike random sampling, this mechanism actively mines difficult examples, such as points near object boundaries or in cluttered regions, where the student's confidence is low. This forces the model to focus on the most challenging features during the primary distillation pass. Finally, the model undergoes post-training quantization \cite{banner2019post, nagel2020up} to minimize memory footprint and inference latency. Despite these optimizations, the student model in TinySAM retains a strong architectural dependence on the transformer-based design of the original SAM. Picosam2 \cite{picosam2} departs from this assumption by adopting a fully custom U-Net-style student. This architecture is trained end-to-end without reusing the pre-defined backbone found in TinySAM, enabling greater flexibility for architectural adjustments and hardware co-design specific to constrained edge devices.

EdgeSAM \cite{zhou2023edgesam} addresses the computational bottleneck of the original SAM by replacing the heavy transformer-based image encoder with a purely CNN-based backbone optimized for edge devices. To prevent performance degradation from this architectural shift, EdgeSAM introduces a \textit{prompt-in-the-loop} distillation strategy. Distinct from simple hard example mining, this method simulates an iterative user interaction session during training: it takes the student's initial prediction, identifies error regions, and feeds corrective prompts back into the model. This forces the lightweight student to learn the temporal dynamics of mask refinement rather than just static segmentation. Similarly to EdgeSAM, our approach selects a region of interest during training and explicitly enables prompt refinement at inference-time.

Beyond static images, SAM2 \cite{ravi2024sam2} extends promptable segmentation to the video domain by incorporating a hierarchical pyramid transformer backbone known as Hiera \cite{ryali2023hiera}. Unlike the original ViT~\cite{dosovitskiy2020image}, Hiera eliminates complex vision-specific components—such as shifted windows or relative position biases—in favor of a streamlined, Masked Autoencoder (MAE) ~\cite{he2022masked} architecture. This design allows SAM2 to process images approximately 6$\times$ faster than its predecessor while managing temporal memory modules for video tracking. Even with these speed gains, the massive parameter count makes it impractical for the tightest edge constraints.

Regarding inference efficiency, MobileSAMv2 \cite{zhang2023mobilesamv2} addresses the computational bottleneck of the \textit{segment everything} task not by changing the network structure, but by introducing object-aware prompt sampling. Instead of querying the decoder with a dense grid of points, it uses a lightweight detector to generate precise box prompts, significantly reducing the number of decoder calls required. Building on this need for efficiency, Picosam2 \cite{picosam2} takes a complementary structural approach: rather than just optimizing prompt quantity, it redesigns the decoder itself. By replacing standard transformer attention mechanisms with CNN-based mask heads, Picosam2 avoids expensive matrix multiplications entirely while preserving spatial precision.

LiteSAM \cite{fu2024litesam} demonstrates that lightweight modular designs can significantly accelerate the \textit{segment everything} task. To achieve this, it replaces standard dense grid prompting with AutoPPN \cite{xu2023auto}, a proposal network that automatically identifies object locations to generate sparse prompts. Additionally, it employs a LiteViT encoder \cite{wang2023litevit} to reduce parameter count. However, despite these reductions, the student model retains a hybrid architecture relying on transformer blocks. This dependence imposes synchronization barriers and unsupported operator overheads (e.g., complex attention mechanisms) that limit deployment on strictly constrained microcontrollers. In contrast, Picosam3 eliminates transformer layers entirely, utilizing a dense CNN backbone. We compensate for the loss of global attention by integrating multi-scale pyramid features \cite{picosam2} and ECA~\cite{wang2020eca}. This combination approximates transformer-like representational capacity while remaining fully compatible with the standard convolutional instruction sets of ultra-low-power in-sensor computing platforms \cite{imx500_sony_sensor, eki2021sonyIMX500}.

Recent work on in-sensor and edge-AI systems~\cite{Kuhne_Resource-ConstrainedUAVs2025, Fan_EfficientRemoteSensing, Campanella_WearableSensor2024, Vitolo_Low-PowerDetection2022} illustrate the broader co-design paradigm in which hardware capabilities and algorithmic choices must be jointly optimised. PicoSAM3 follows this principle: the CNN-only architecture satisfies the IMX500 op-set constraint, the INT8 quantisation meets the 2\,MB SRAM limit, and the ROI-as-crop strategy leverages the sensor's native hardware ROI API.

Overall, PicoSAM3 builds upon the structural efficiency of PicoSAM2 \cite{picosam2} but introduces critical advancements to address its limitations. We generalize the restrictive centered-point prompting to flexible ROI-based supervision and upgrade the knowledge distillation process by learning directly from the semantically superior SAM 3 \cite{carion2025sam3segmentconcepts}. Collectively, the proposed contributions significantly improve spatial flexibility and segmentation quality while adhering to the strict power and memory constraints required for in-sensor deployment.

For a complete overview of the results, please refer to the Table~\ref{tab:merged_results}.
\input{tables/related}

\section{Hardware Setup}

\begin{figure}[h]
    \includegraphics[width=\linewidth]{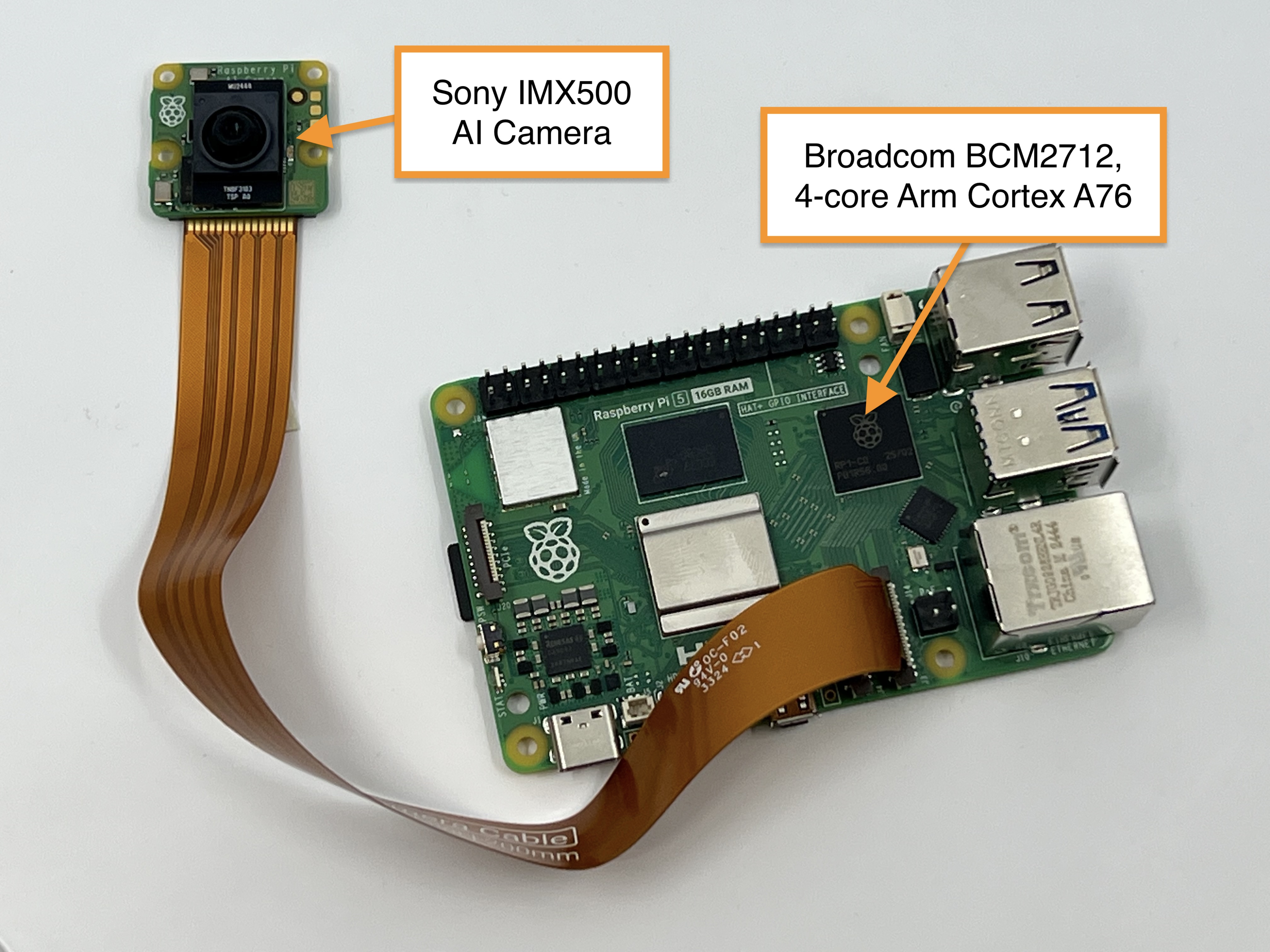} 
    \caption{Hardware setup: PicoSAM3 runs directly in the Sony IMX500 camera, the ARM Cortex A76 is only used to deploy the model.}
    \label{fig:hardware_setup}
\end{figure}

The hardware setup consists of a Raspberry Pi 5 connected to a Sony IMX500 \cite{imx500_sony_sensor} intelligent vision sensor via the MIPI CSI-2 interface, illustrated in Figure~\ref{fig:hardware_setup}. 

The Sony IMX500 sensor \cite{eki2021sonyIMX500, imx500_sony_sensor} enables vision models to run directly in-sensor, reducing system latency and energy consumption \cite{bonazzi2023tinytracker}. It showed edge vision state-of-the-art efficiency performance in gaze estimation \cite{bonazzi2023tinytracker}, anomaly detection \cite{bonazzi2026tinyglass}, health monitoring \cite{Tong2024} and smart cities \cite{Cui2024}. 

Like its predecessor \cite{picosam2}, PicoSAM3 runs fully in-sensors on the Sony IMX500, and Rasperry Pi 5 is used only to deploy the model on the camera.

\section{PicoSAM3}
\label{sec:metodology}

\subsection{Implicit Prompt Encoding via Centered Cropping Square-Crop Filter Chain}

Due to the strict constraint of RGB-only inputs on the IMX500, explicit prompt representations (e.g., point coordinates, box tensors, or mask channels) cannot be provided as additional network inputs. To enable prompt-conditioned segmentation under this limitation, we adopt an implicit prompting strategy that encodes the spatial prompt through centered cropping. During training, given an input image of dimensions $(W, H)$ and an object instance with COCO-format bounding box $[x, y, w_b, h_b]$ (top-left corner and dimensions), we extract a square crop centered on the object. The bounding box is extended by a padding factor $p = 0.1$ to include contextual information around the object:

\begin{align}                                                                           w' &= w_b \cdot (1 + 2p), \quad h' = h_b \cdot (1 + 2p)
\end{align}

To ensure a consistent aspect ratio, we compute the crop size as the maximum of the padded dimensions:                                                                    
  \begin{equation}
  s = \max(w', h')                             
  \end{equation}
  
  Next, the crop is centered on the bounding box center $(c_x, c_y)$:                 
  \begin{equation}
  c_x = x + \frac{w_b}{2}, \quad c_y = y + \frac{h_b}{2}
  \end{equation}                                   
  
  \begin{equation}
    \begin{aligned}
  x_1 = c_x - \frac{s}{2}, \quad y_1 = c_y - \frac{s}{2}, \\
  \quad x_2 = x_1 + s, \quad y_2 = y_1 + s 
   \end{aligned}
  \end{equation}
  The crop coordinates are then clamped to remain within valid image bounds: 
  \begin{equation}
  \begin{aligned}
  x_1 = \max(0, x_1), 
  \quad y_1 = \max(0, y_1), \\
  \quad x_2 = \min(W, x_2), 
  \quad y_2 = \min(H, y_2) 
  \end{aligned}
  \end{equation}

The resulting crop region $(x_1, y_1, x_2, y_2)$ is extracted from both the RGB image and the ground truth mask, then resized to the fixed network input resolution $S \times S$ (where $S = 96$ in our implementation) using bilinear interpolation for images and nearest-neighbor interpolation for masks.                               
This cropping strategy implicitly encodes the bounding box prompt through spatial normalization: the target object is always approximately centered in the input crop, and the crop boundaries define the spatial extent of interest. The network learns to segment the dominant object near the image center, effectively treating the crop center as an implicit point prompt and the crop boundaries as an implicit box prompt. The design also enables seamless deployment on the IMX500 sensor, through the sensor ROI function.    

\subsection{Model Architecture}

The proposed architecture is inspired by its predecessor, PicoSAM2 \cite{picosam2}. PicoSAM2 is a a symmetric encoder-decoder U-Net~\cite{ronneberger2015unet} with four stages featuring depthwise separable convolutions~\cite{chollet2017xception}.  


The encoder progressively increases channel dimensions (48→96→160→256) using depthwise separable convolutions, each followed by strided convolution for downsampling. A bottleneck layer expands to 320 channels before the decoder mirrors this structure with nearest-neighbor upsampling and skip connections from corresponding encoder stages. The final 1×1 convolution produces the segmentation logit. This design yields a model of only 1.26M parameters (4.84 MB), reducible to 1.21 MB with INT8 quantization. 
  
Building upon PicoSAM2, we introduce three architectural enhancements: (1) an \textit{enhanced bottleneck} with dilated depthwise convolution (dilation=2) to expand the receptive field without additional downsampling, (2) an \textit{Efficient Channel Attention (ECA)} block~\cite{wang2020eca} before the output head to enable adaptive feature recalibration, and (3) a \textit{refinement head} with depthwise convolution for improved boundary delineation. These additions increase the model to 1.37M parameters (5.26 MB), or 1.31 MB quantized—an 8.6\% overhead for improved segmentation quality. 

Rather than encoding prompts explicitly, we leverage the ground truth bounding box annotations to define regions of interest. For each annotated object instance, we extract the bounding box, apply 10\% padding on each side to include contextual information, and convert it to a square crop by taking the maximum of width and height. This square region is then cropped from both the image and segmentation mask, and resized to the target resolution ($96 \times 96$). 

During knowledge distillation, the teacher model (SAM2 or SAM3) receives the same bounding box as a geometric prompt and produces soft mask logits, which are cached for efficient training. The student network observes only the cropped RGB image—with no explicit prompt encoding—and learns to predict segmentation masks that match the teacher's output. Since each crop is inherently centered on the object of interest, the network implicitly learns that the target object lies at the center of its input, eliminating the need for additional prompt channels while maintaining promptability through the crop-based spatial prior.

This approach offers two key advantages: (1) it naturally aligns with how bounding box detectors operate in practical pipelines, enabling seamless integration where a detector provides boxes and PicoSAM refines them into precise masks, and (2) it simplifies the input representation to pure RGB, facilitating deployment on resource-constrained hardware without auxiliary prompt encoding modules.
  
\subsection{SAM3 Distillation}

We employ a two-stage knowledge distillation pipeline to transfer segmentation capabilities from the large SAM3 teacher \cite{carion2025sam3segmentconcepts} (1.2 GB) to our compact PicoSAM3 student (5.26 MB). To enable efficient training, we precompute teacher predictions for all COCO training annotations. For each annotation, SAM3 receives the corresponding bounding box as a geometric prompt and generates a soft probability mask. We extract the region of interest around the bounding box with 10\% padding, resize it to $96 \times 96$ pixels, and store both the logits and the teacher's confidence score. This offline caching strategy eliminates the need to run the expensive teacher model during training, reducing GPU memory requirements and enabling larger batch sizes.               

Our training objective combines three complementary loss terms. 

To transfer dark knowledge from the teacher's soft predictions \cite{hinton2015distilling}, we use a combination of Mean Squared Error (MSE) and Dice loss~\cite{milletari2016v}: 

\begin{align}                                
\mathcal{L}_{\text{teacher}} &= \text{MSE}(\sigma_\tau(\hat{y}), \sigma_\tau(y_t)) + \mathcal{L}_{\text{Dice}}(\sigma_\tau(\hat{y}), \sigma_\tau(y_t))                                    
\end{align}                                  
where $\hat{y}$ denotes the student's logits, $y_t$ the teacher's logits, and $\sigma_\tau(x) = \sigma(\tau \cdot x)$ is a temperature-scaled sigmoid with $\tau = 5$ to produce sharper  
probability distributions. The MSE term encourages the student to match the teacher's soft boundaries, while the Dice term \cite{milletari2016vnet} ensures overlap consistency.          

For hard supervision from binary masks, we employ balanced Binary Cross Entropy (BCE) combined with Dice loss:
\begin{align}                                
\mathcal{L}_{\text{gt}} &= \text{BCE}(\sigma_\tau(\hat{y}), y_{\text{gt}}) + \mathcal{L}_{\text{Dice}}(\sigma_\tau(\hat{y}), y_{\text{gt}})                                               
\end{align}                                  
where $y_{\text{gt}}$ is the ground truth binary mask. This dual formulation addresses the class imbalance between foreground and background pixels common in segmentation tasks          
\cite{ronneberger2015unet}.

To prevent model collapse where the student predicts trivially small masks, we introduce an area preservation term:                                  
  \begin{align}                                
  \mathcal{L}_{\text{area}} &= \max\left(0, \rho - \frac{\sum \sigma_\tau(\hat{y})}{\sum y_{\text{gt}}}\right)                                                                              
  \end{align}                                  
  where $\rho = 0.4$ is a minimum area ratio threshold. This loss activates only when the predicted mask area falls below 40\% of the ground truth area.

The total training loss adaptively balances teacher and ground truth supervision based on the teacher's confidence:                                                                       
\begin{align}                                
\mathcal{L}_{\text{total}} &= \alpha \cdot \mathcal{L}_{\text{teacher}} + (1 - \alpha) \cdot \mathcal{L}_{\text{gt}} + 0.4 \cdot \mathcal{L}_{\text{area}}                                
\end{align}                                  
where $\alpha \in [0, 1]$ is set to the mean teacher confidence score for each batch, clamped to $[0, 1]$. When the teacher is highly confident (high $\alpha$), the student relies more  
on the teacher's soft predictions; when confidence is low, the student defaults to ground truth supervision. This adaptive weighting scheme, inspired by curriculum learning principles   
\cite{Sanh2019DistilBERTAD}, allows the student to learn from reliable teacher predictions while avoiding potentially erroneous guidance.   

\begin{table*}[!t]
\centering
\caption{Comparison of edge segmentation models. GPU latency measured on NVIDIA T4, DPU latency on Sony IMX500 (when available).}
\label{tab:merged_results}
\setlength{\tabcolsep}{6pt}
\begin{tabular*}{\textwidth}{
@{\extracolsep{\fill}}
l c r r c c r r r}
\toprule
 &  &
\multicolumn{2}{c}{\textbf{Complexity$\downarrow$}} &
\multicolumn{2}{c}{\textbf{mIoU (\%) $\uparrow$}} & &
\multicolumn{2}{c}{\textbf{Latency (ms)$\downarrow$}} \\
\cmidrule(lr){3-4}
\cmidrule(lr){5-6}
\cmidrule(lr){8-9}
\textbf{Model} &
\textbf{Prompt} &
\textbf{Params (M)} &
\textbf{MACs} &
\textbf{COCO} &
\textbf{LVIS} &
\textbf{Size (MB)} &
\textbf{GPU} &
\textbf{IMX500} \\
\midrule
SAM-H~\cite{kirillov2023segment}        & -- & 635  & 2.97T & 53.6  & 60.5  & 2420.3 & 2390  & -- \\
FastSAM~\cite{zhao2023fast}     & -- & 72.2 & 443G  & 51.6  & 55.2  & 275.6  & 153.6 & -- \\
TinySAM~\cite{shu2023tinysam}     & -- & 9.7  & 42G   & 50.9  & 52.1  & 37.0   & 38.4  & -- \\
EdgeSAM~\cite{zhou2023edgesam}     & -- & 9.7  & 42G   & 48.0  & 53.7  & 37.0   & 38.4  & -- \\
LiteSAM~\cite{fu2024litesam}     & -- & 4.2  & 4.2G  & 49.0  & 51.2  & 16.0   & 16.0  & -- \\
Supervised~\cite{picosam2}  & -- & 1.3  & 336M  & 53.0  & 41.4  & 4.87   & 2.5   & -- \\
PicoSAM2~\cite{picosam2}    & Center & 1.3  & 336M  & 51.9  & 44.9  & 4.87   & 2.54  & -- \\
Q-PicoSAM2~\cite{picosam2}  & Center & 1.3  & 324M  & 50.5  & 45.1  & 1.22   & --    & 14.3 \\  \midrule
PicoSAM3    & Box    & 1.37 & 345M  & 65.45 & 64.01 & 5.26 & -- & -- \\
Q-PicoSAM3  & Box    & 1.37 & 345M  & 65.34 & 63.98 & 1.30   & --    & 11.82 \\
\bottomrule
\end{tabular*}
\end{table*}

\subsection{Training and Inference Pipeline}

We train PicoSAM3 using the AdamW optimizer \cite{loshchilov2019decoupledweightdecayregularization} with a learning rate of $3 \times 10^{-4}$ and a linear warmup schedule over the first 1000 steps. Training is 
performed on COCO \cite{lin2014microsoft}  with a batch size of 64 for one epoch. 

After training, we apply post-training quantization (PTQ) to INT8 precision using Sony's Model Compression Toolkit (MCT) \cite{habi2021hptq, gordon2024eptq, dikstein2025dgh}. The quantization pipeline proceeds as follows: (1) load the trained floating-point model; (2) calibrate activation ranges using a representative dataset of 10 batches sampled from COCO val2017; (3) apply symmetric per-channel quantization for weights and per-tensor quantization for activations; (4) export the quantized model to ONNX format for hardware compilation. The resulting INT8 model achieves a 4$\times$ reduction in size (from 5.26 MB to 1.31 MB) while preserving segmentation accuracy, as the depthwise separable convolution architecture exhibits inherent robustness to quantization noise \cite{gholami2021surveyquantizationmethodsefficient}.    

For on-device deployment, we target the Sony IMX500 intelligent vision sensor \cite{eki2021sonyIMX500, imx500_sony_sensor}, a stacked back-side illuminated (BSI) CMOS image sensor with an integrated digital signal processor optimized for CNN inference. The sensor captures images at $4056 \times 3040$ resolution and features 2304 multiply-accumulate (MAC) units operating at 262.5 MHz, delivering up to 4.97 TOPS/W energy efficiency.                  
The quantized ONNX model is compiled to the IMX500's \texttt{.rpk} format and deployed on-sensor. Our deployment implements interactive promptable segmentation through the sensor's hardware ROI capability. Users provide bounding box prompts by dragging a rectangle on the display interface; the coordinates are transformed from display space ($640 \times 480$) to sensor space ($4056 \times 3040$) and passed to the IMX500 via the \texttt{set\_inference\_roi\_abs()} API. The sensor then crops the specified region, resizes it to the model's input resolution ($96 \times 96$), executes inference, and returns the segmentation mask.                                  

The inference pipeline operates asynchronously: a callback function receives completed requests containing both the camera frame and CNN outputs. The output mask is thresholded ($> 0$), resized to match the ROI dimensions, and composited onto the display frame as a colored overlay. ROI updates are rate-limited to 150ms intervals to ensure stable inference. 

This deployment demonstrates that promptable segmentation, previously requiring desktop GPUs with multi-gigabyte models, can operate on a sub-watt intelligent sensor with 1.31 MB of storage. The bounding box prompting paradigm maps naturally to the IMX500's ROI inference mode, enabling practical applications in embedded robotics, wearable AR devices, and battery-powered IoT systems where power consumption, latency, and form factor are critical constraints.

\section{Experimental Results}
\label{sec:experimental_results}

This section reports the quantitative results on COCO and LVIS, focusing on the trade-off between accuracy, model size, and deployment latency. All training was conducted on a workstation equipped with an NVIDIA RTXA6000-48Q GPU while we used a Tesla T4 GPU for evaluation. The models were implemented using PyTorch 3.7 with CUDA 12.6. To assess real-world performance on edge devices, we utilized the hardware configuration illustrated in Figure~\ref{fig:hardware_setup}. The inference runs fully in-sensors on the IMX500, and the RPI is used only to deploy the code on the camera. 

\subsection{Model Performance}

Table~\ref{tab:merged_results} provides a comprehensive comparison of the PicoSAM variants against state-of-the-art foundation models and compact edge-oriented baselines.

While large-scale models like SAM-H establish a strong baseline, their computational cost is prohibitive for edge constraints, requiring 635M parameters and 2.97T MACs. Remarkably, PicoSAM3 not only reduces the parameter count by over 460 times (1.37M vs. 635M) but also surpasses SAM-H in segmentation accuracy, achieving a gain of +11.85 mIoU on COCO (65.45\% vs. 53.6\%) and +3.5 mIoU on LVIS. Similarly, compared to FastSAM, PicoSAM3 delivers a +13.85\% improvement in COCO mIoU while requiring only 1.9\% of the model size (5.26 MB vs. 275.6 MB).

\begin{figure}[h]
\centering
\includegraphics[width=\linewidth]{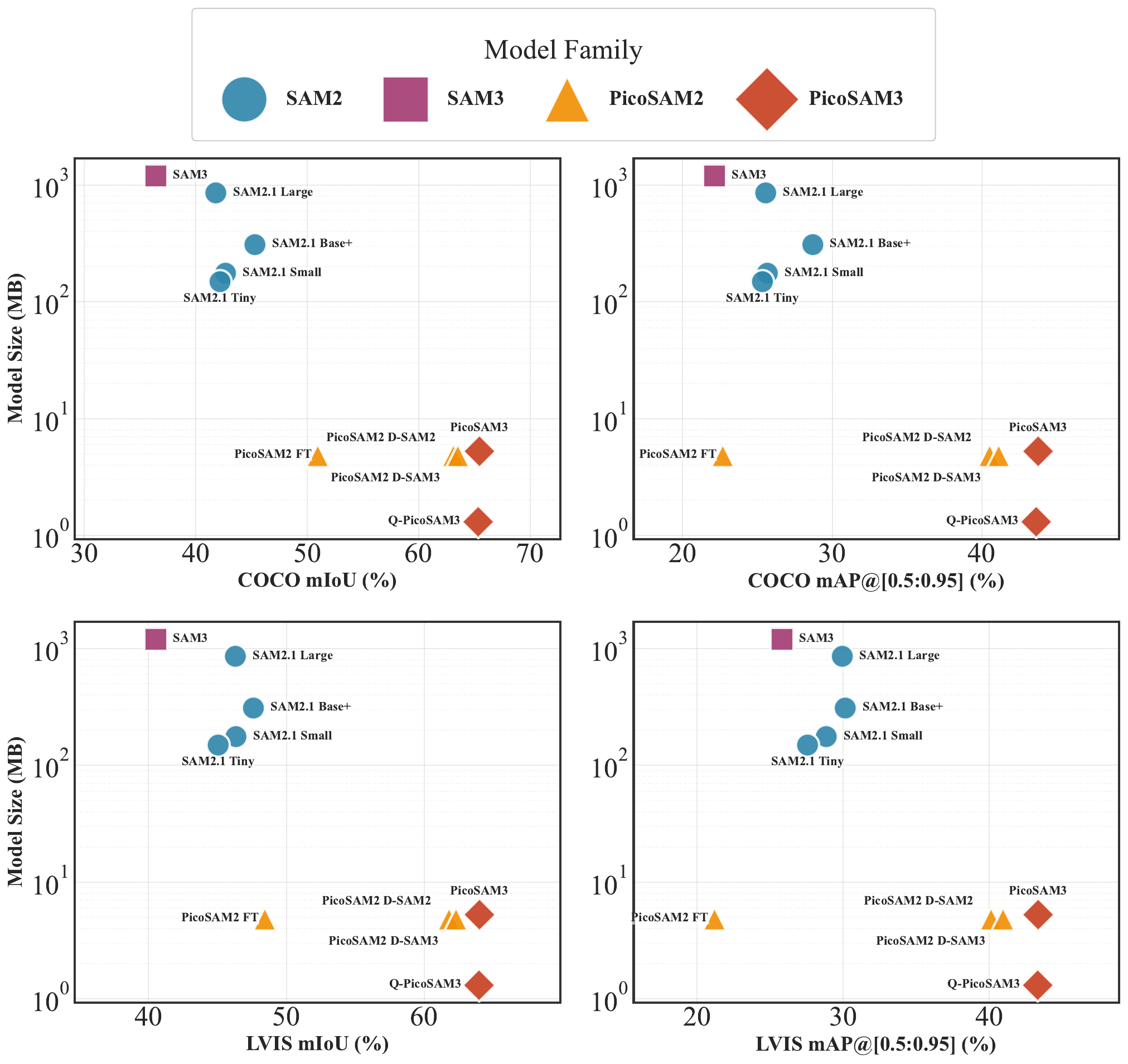}
\caption{Segmentation precision (mAP and mIoU) versus model size (log scale) across model families.}
\label{fig:performance_picosam3}
\end{figure}

Figure~\ref{fig:performance_picosam3} further highlights PicoSAM3's strong efficiency-accuracy trade-off. Large foundation models suffer severe degradation at this resolution due to their high-resolution bias, whereas PicoSAM3 is specifically optimized for the $96\times96$ regime required by in-sensor computing.

Against other lightweight architectures, PicoSAM3 demonstrates superior efficiency-accuracy trade-offs. It outperforms TinySAM and EdgeSAM by substantial margins, exceeding their COCO performance by over 14 mIoU while maintaining a significantly lower computational footprint (345M MACs vs. 42G MACs). This confirms that PicoSAM3 effectively distills the segmentation capability of larger models better than existing pruning or distillation techniques.

The transition from PicoSAM2 to PicoSAM3 yields the most significant performance leap. PicoSAM3 improves upon its predecessor by +13.55 mIoU on COCO and a remarkable +19.11 mIoU on the LVIS dataset. This enhancement highlights the efficacy of the updated prompting strategy and architectural refinements.

Finally, the quantization results validate the feasibility of real-time edge execution. Q-PicoSAM3 incurs a negligible accuracy loss of just 0.11\% (COCO) compared to its floating-point counterpart. Crucially, when deployed on the Sony IMX500 DSP, Q-PicoSAM3 achieves a latency of 11.82 ms. This represents a 17\% speedup over Q-PicoSAM2 (14.3 ms), thereby ensuring true real-time performance (approx. 84 FPS) for end-to-end inference.

\subsection{System Performance}

PicoSAM3 operates in two deployment modes. In \emph{interactive mode}, the user specifies a bounding box via the companion display; no detector is needed, and the only latency is the 11.82\,ms segmentation inference. In \emph{cascaded mode}, a lightweight detector provides boxes autonomously. Since the IMX500 budget accommodates a second sub-1.5\,MB model, the total pipeline latency is $\approx$\,15\,ms (detection) $+$\,11.82\,ms (segmentation) $\approx$\,27\,ms (37\,FPS). Our implementation rate-limits ROI updates to 150\,ms to ensure stable, jitter-free inference while the segmenter runs at full 84\,FPS.

Using the IMX500's rated efficiency of 4.97\,TOPS/W~\cite{eki2021sonyIMX500}, and given 690\,MOPS per inference (2\,$\times$\,345\,MMAC) at 11.82\,ms, the effective compute throughput is $\approx$\,58.4\,GOPS, yielding an estimated power of $P \approx 58.4\,\text{GOPS}\,/\,4.97\,\text{TOPS/W} \approx 11.75\,\text{mW}$ and energy per inference of $E \approx 11.75\,\text{mW} \times 11.82\,\text{ms} \approx 139\,\mu\text{J}$. This is two to three orders of magnitude below a GPU inference and is consistent with battery-operated IoT operation.

\subsection{Ablation Studies}

We conduct extensive ablations to quantify the contribution of each design choice in PicoSAM3. 

\subsubsection{Architectural Components}

Table~\ref{tab:ablations_prompt} systematically evaluates the contribution of ROI prompting, knowledge distillation, and architectural improvements. 

\input{tables/ablations.tex}

The transition to **ROI-based inputs** brings the largest single gain, improving mAP from 29.96\% to 40.52\% under SAM2 distillation on COCO. This highlights the importance of explicit spatial conditioning for low-resolution inference. 

Knowledge distillation consistently outperforms supervised training, with SAM3 providing the strongest teacher. The two components are synergistic: distillation alone yields only +0.9\% mIoU over the supervised baseline, while ROI alone slightly hurts performance without distillation. Together they deliver a +10.1\% mIoU gain. 

Our final architectural enhancements (dilated bottleneck and Efficient Channel Attention) contribute an additional +1.9\% mIoU. In total, PicoSAM3 surpasses the best PicoSAM2 configuration by +2.6\% mAP / +2.3\% mIoU and sets a new state-of-the-art for this constraint class. The INT8 quantized model (Q-PicoSAM3) maintains near-lossless accuracy with a $4\times$ reduction in size (1.31\,MB).

\subsubsection{Prompt Sensitivity}

Table~\ref{tab:padding_ablation} evaluates the effect of context padding ratio $p$ (trained at $p=10\%$). Performance peaks at $p=10\%$ on both datasets (65.3\% mIoU on COCO and 63.7\% mIoU on LVIS). Zero padding ($p=0\%$) causes boundary information loss (56.9\%/56.1\% mIoU), while excessive padding ($p=20\%$) introduces background clutter and degrades results to 55.7\%/54.5\%. This confirms that $p=10\%$ offers the optimal trade-off between context and focus.

\input{tables/padding_ablation.tex}

We also evaluate robustness to noisy bounding box prompts, which is critical for practical cascaded systems. As shown in Table~\ref{tab:spatial_robustness}, PicoSAM3 exhibits graceful degradation under center shifts and scale perturbations, maintaining competitive performance even at moderate noise levels.

\begin{table}[h]
  \centering\small
  \caption{PicoSAM3 robustness to spatial bounding box perturbations (mean $\pm$ std over 5 trials). Each column shows a perturbation level expressed as a fraction of the ground-truth bbox size (uniform center shift $\pm s/2$, Gaussian dimension noise $\sigma = s/2$).}
  \label{tab:spatial_robustness}
  \setlength{\tabcolsep}{3.8pt}
  \resizebox{\columnwidth}{!}{%
    \begin{tabular}{llcccccccc}
      \toprule
      \textbf{Dataset} & \multicolumn{2}{c}{\textbf{0\%}} & \multicolumn{2}{c}{\textbf{5\%}} & \multicolumn{2}{c}{\textbf{10\%}} & \multicolumn{2}{c}{\textbf{20\%}} \\
      \cmidrule(lr){2-3} \cmidrule(lr){4-5} \cmidrule(lr){6-7} \cmidrule(lr){8-9}
      & mIoU & mAP & mIoU & mAP & mIoU & mAP & mIoU & mAP \\
      \midrule
      COCO & 0.653 & 0.435 & 0.644 & 0.422 & 0.625 & 0.394 & 0.581 & 0.330 \\
      LVIS & 0.637 & 0.429 & 0.629 & 0.417 & 0.611 & 0.391 & 0.569 & 0.329 \\
      \bottomrule
    \end{tabular}%
  }
\end{table}

\subsubsection{Multi-Object Segmentation}

Table~\ref{tab:multiobj_robustness} evaluates PicoSAM3 segmentation performance stratified by number of competing objects whose centre falls inside the padded bounding-box crop. Dense crops (3\,+ competing objects) drop mIoU performance by ${\approx}$\,14\,points relative to unambiguous crops.

\begin{table}[b]
  \centering
  \small
  \caption{Segmentation performance stratified by number of competing objects whose centre falls inside the padded bounding-box crop.}
  \label{tab:multiobj_robustness}
  \setlength{\tabcolsep}{5pt}
  \begin{tabular}{lcccccc}
    \toprule
    \textbf{Dataset} & \multicolumn{2}{c}{\textbf{0 competitors}} & \multicolumn{2}{c}{\textbf{1-2 competitors}} & \multicolumn{2}{c}{\textbf{3+ competitors}} \\
    \cmidrule(lr){2-3} \cmidrule(lr){4-5} \cmidrule(lr){6-7}
    & mIoU & mAP & mIoU & mAP & mIoU & mAP \\
    \midrule
    COCO & 0.706 & 0.504 & 0.637 & 0.411 & 0.565 & 0.329 \\
    LVIS & 0.682 & 0.478 & 0.638 & 0.432 & 0.521 & 0.299 \\
    \bottomrule
  \end{tabular}
\end{table}

  \begin{figure*}[h]                          
      \centering                          
      \includegraphics[width=\textwidth]{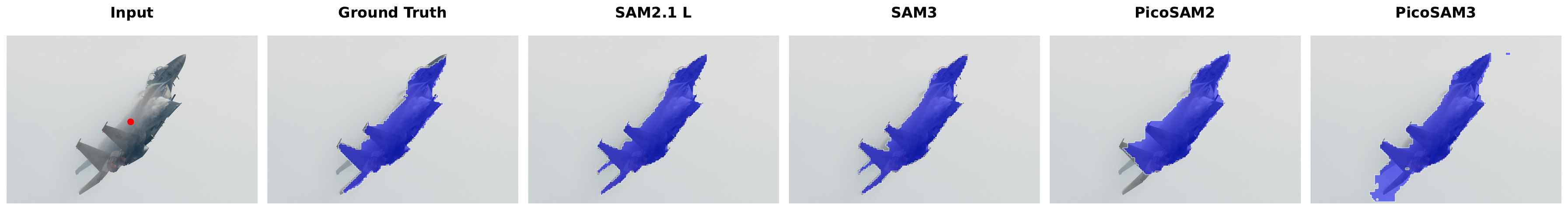}\\
      \includegraphics[width=\textwidth]{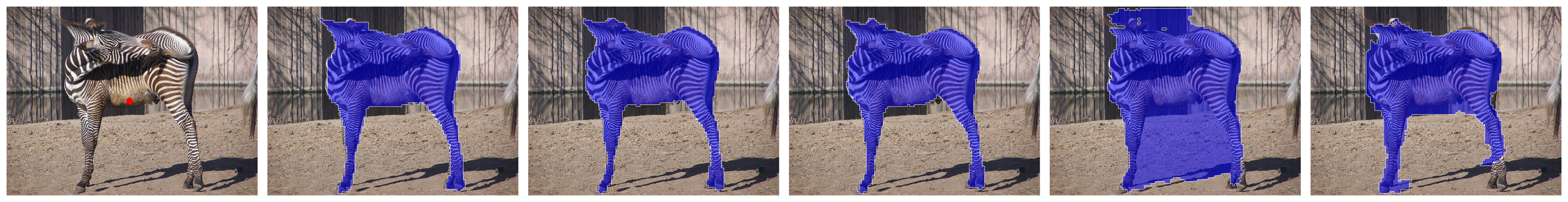}\\
      \includegraphics[width=\textwidth]{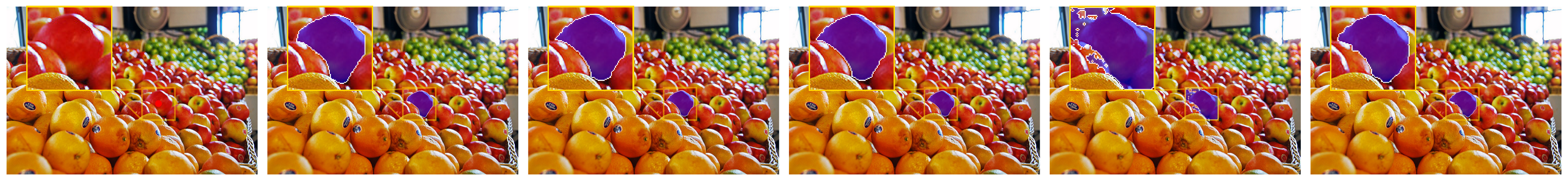}\\    
      \includegraphics[width=\textwidth]{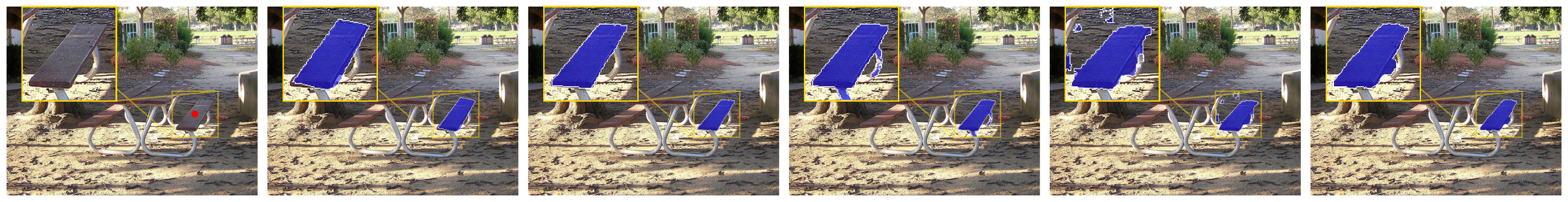}\\
      \caption{Qualitative comparison of each model's mask inference.}                                                               
      \label{fig:qualitative_banner}                                                                                                  
  \end{figure*}  

\subsection{Qualitative Results}

Figure~\ref{fig:qualitative_banner} presents a qualitative comparison between the proposed PicoSAM3 and, PicoSAM2, alongside the foundation models SAM2.1 Large and SAM3. As observed, PicoSAM3 outperforms PicoSAM2, exhibiting sharper boundary adherence and a marked reduction in segmentation artifacts. Moreover, as exemplified in the first row, PicoSAM3 robustly disambiguates cases where the SAM3 teacher struggles to distinguish the object in the foreground. Notably, the visual outputs of PicoSAM3 demonstrate a substantial degree of perceptual alignment with the heavy-weight SAM2 and SAM3 baselines, confirming that the distilled model retains the structural integrity of the foundation models while operating at a fraction of the computational cost. 

Despite strong average performance, PicoSAM3 exhibits four principal failure modes. (1)~\emph{Transparent and textureless objects} (glass, water, mirrors): when foreground and background share the same texture statistics, the model cannot distinguish the target region. (2)~\emph{Severely occluded instances}: when less than ${\approx}\,30$\% of the annotated object is visible in the crop, the remaining evidence is insufficient for reliable mask estimation. (3)~\emph{Sub-resolution small objects} ($<$\,10$\,\times\,$10\,px at 96$\,\times\,$96): the model over-segments toward the full crop area when the discriminative region is too small to produce meaningful features. (4)~\emph{Amorphous objects} (clouds, smoke, liquid pours): the lack of clear boundaries produces diffuse, unreliable masks even when the bounding box is accurate. These failure modes are consistent with those observed in larger SAM models \cite{zhang2023survey} and are exacerbated by the low input resolution. 

\section{Conclusion and Discussion}
\label{sec:conclusion}

This work presents PicoSAM3, an ultra-lightweight (1.3--1.4\,M parameters) promptable segmentation model optimized for real-time in-sensor deployment. By combining a dense CNN architecture with ROI-based prompting, Efficient Channel Attention, and knowledge distillation from SAM3, PicoSAM3 achieves state-of-the-art results among edge-oriented models: 65.45\% mIoU on COCO and 64.01\% mIoU on LVIS, while substantially outperforming prior SAM-based and lightweight baselines at similar or lower complexity.

The INT8 quantized model incurs negligible accuracy loss ($<0.2\%$ mIoU) and enables real-time inference at 11.82\,ms on the Sony IMX500, fully respecting its strict memory and operator constraints. Ablation studies confirm that distillation from SAM3 yields up to +14.5\% mIoU over supervised training, while ROI prompting significantly improves spatial conditioning compared to centered-point supervision. The final architectural enhancements further boost performance without increasing latency or memory footprint.

PicoSAM3 demonstrates that high-quality, spatially flexible promptable segmentation is feasible directly at the sensor level. This is enabled by tight hardware-software co-design across three axes: operator compatibility (CNN-only backbone), memory footprint (tuned channel widths for 1.31\,MB INT8), and prompt integration (native ROI cropping via the IMX500 API).These results open a promising path toward scalable, privacy-preserving, and low-latency vision intelligence in next-generation edge and in-sensor systems.

\section{Reproducibility }
\label{sec:reproducibility}

The codebase to replicate the experimental results and to deploy the model on the IMX500 is publicly available at: \url{https://github.com/pbonazzi/picosam3}.

\bibliographystyle{IEEEtran}
\bibliography{main}


\end{document}

%% file: tables/related.tex






%% file: tables/ablations.tex
\begin{table}[htbp]

    \centering

    \caption{Ablation results for region of interest cropped inputs across datasets, distillation teachers and architectures.}

    \label{tab:ablations_prompt}
    \resizebox{0.48\textwidth}{!}{
    \begin{tabular}{lcccccc}

        \toprule

        \multirow{2}{*}[-0.8mm]{\textbf{Model}} & \multirow{2}{*}[-0.8mm]{\textbf{Distillation}} & \multirow{2}{*}[-0.8mm]{\textbf{ROI}} & \multicolumn{2}{c}{\textbf{COCO}} & \multicolumn{2}{c}{\textbf{LVIS}} \\
        \cmidrule(lr){6-7}
        \cmidrule(lr){4-5}
        & & & \textbf{mAP} & \textbf{mIoU} & \textbf{mAP} & \textbf{mIoU} \\
        
        \midrule
        \multirow{5}{*}{PicoSAM2~\cite{picosam2}} & \myxmark & \myxmark & 30.67 & 53.00 & 20.38 & 41.40\\
                                                  & \myxmark & \mycheckmark & 22.68 & 50.94 & 21.20 & 48.44\\
                                                  & SAM2\cite{ravi2024sam2} &\myxmark & 29.96 & 51.93 & 25.46 & 44.88\\
                                                  & SAM2\cite{ravi2024sam2} & \mycheckmark & 40.52 & 63.11 & 40.12 & 61.80\\
                                                  & SAM3\cite{carion2025sam3segmentconcepts} & \mycheckmark & 41.13 & 63.51 & 40.94 & 62.31\\
        \midrule
        PicoSAM3 & SAM3\cite{carion2025sam3segmentconcepts} & \mycheckmark & \textbf{43.77} & \textbf{65.45} & \textbf{43.34} & \textbf{64.01}\\                                  
        Q-PicoSAM3 & SAM3\cite{carion2025sam3segmentconcepts} & \mycheckmark & 43.64 & 65.34 & 43.31 & 63.98\\                                  

        \bottomrule
    \end{tabular}
    }
\end{table} 

%% file: tables/padding_ablation.tex
\begin{table}[ht]
  \centering\small
  \caption{Effect of ROI padding percentage $p$ on inference-time performance (model trained with $p\!=\!10\%$). Results confirm $p\!=\!10\%$ is the optimal context window.}
  \label{tab:padding_ablation}
  \setlength{\tabcolsep}{3.5pt}
  \resizebox{\columnwidth}{!}{%
    \begin{tabular}{lcccccccccc}
      \toprule
      \multirow{2}{*}[-0.8mm]{\textbf{Dataset}} &
      \multicolumn{2}{c}{\textbf{0\%}} &
      \multicolumn{2}{c}{\textbf{5\%}} &
      \multicolumn{2}{c}{\textbf{10\%}} &
      \multicolumn{2}{c}{\textbf{15\%}} &
      \multicolumn{2}{c}{\textbf{20\%}} \\
      \cmidrule(lr){2-3}\cmidrule(lr){4-5}\cmidrule(lr){6-7}\cmidrule(lr){8-9}\cmidrule(lr){10-11}
      & mIoU & mAP & mIoU & mAP & mIoU & mAP & mIoU & mAP & mIoU & mAP \\
      \midrule
      COCO & 0.569 & 0.302 & 0.627 & 0.395 & \textbf{0.653} & \textbf{0.435} & 0.613 & 0.380 & 0.557 & 0.304 \\
      LVIS & 0.561 & 0.311 & 0.616 & 0.398 & \textbf{0.637} & \textbf{0.429} & 0.599 & 0.375 & 0.545 & 0.302 \\
      \bottomrule
    \end{tabular}%
  }
\end{table}

%% file: main.bbl
\begin{thebibliography}{10}
\providecommand{\url}[1]{#1}
\csname url@samestyle\endcsname
\providecommand{\newblock}{\relax}
\providecommand{\bibinfo}[2]{#2}
\providecommand{\BIBentrySTDinterwordspacing}{\spaceskip=0pt\relax}
\providecommand{\BIBentryALTinterwordstretchfactor}{4}
\providecommand{\BIBentryALTinterwordspacing}{\spaceskip=\fontdimen2\font plus
\BIBentryALTinterwordstretchfactor\fontdimen3\font minus \fontdimen4\font\relax}
\providecommand{\BIBforeignlanguage}[2]{{%
\expandafter\ifx\csname l@#1\endcsname\relax
\typeout{** WARNING: IEEEtran.bst: No hyphenation pattern has been}%
\typeout{** loaded for the language `#1'. Using the pattern for}%
\typeout{** the default language instead.}%
\else
\language=\csname l@#1\endcsname
\fi
#2}}
\providecommand{\BIBdecl}{\relax}
\BIBdecl

\bibitem{kirillov2023segment}
A.~Kirillov, E.~Mintun, N.~Ravi, H.~Mao, C.~Rolland, L.~Gustafson, T.~Xiao, S.~Whitehead, A.~C. Berg, W.-Y. Lo, P.~Dollár, and R.~Girshick, ``Segment anything,'' \emph{IEEE/CVF International Conference on Computer Vision (ICCV)}, 2023.

\bibitem{ravi2024sam2}
N.~Ravi, V.~Gabeur, Y.~Hu, R.~Hu, C.~Ryali, T.~Ma, H.~Khedr, R.~Rädle, C.~Rolland, L.~Gustafson, E.~Mintun, J.~Pan, K.~V. Alwala, N.~Carion, C.~Wu, R.~Girshick, P.~Dollár, and C.~Feichtenhofer, ``Sam2: Segment anything in images and videos,'' \emph{arXiv, 2408.00714}, 2024.

\bibitem{carion2025sam3segmentconcepts}
N.~Carion, L.~Gustafson, Y.-T. Hu, S.~Debnath, R.~Hu, D.~Suris, C.~Ryali, K.~V. Alwala, H.~Khedr, A.~Huang, J.~Lei, T.~Ma, B.~Guo, A.~Kalla, M.~Marks, J.~Greer, M.~Wang, P.~Sun, R.~Rädle, T.~Afouras, E.~Mavroudi, K.~Xu, T.-H. Wu, Y.~Zhou, L.~Momeni, R.~Hazra, S.~Ding, S.~Vaze, F.~Porcher, F.~Li, S.~Li, A.~Kamath, H.~K. Cheng, P.~Dollár, N.~Ravi, K.~Saenko, P.~Zhang, and C.~Feichtenhofer, ``Sam 3: Segment anything with concepts,'' \emph{arXiv, 2511.16719}, 2025.

\bibitem{Fan_EfficientRemoteSensing}
F.~Fan, M.~Zhang, D.~Yu, J.~Li, and G.~Liu, ``Efficient remote sensing image target detection network with shape-location awareness enhancements,'' \emph{IEEE Sensors Journal}, 2024.

\bibitem{picosam2}
P.~Bonazzi, N.~Farronato, S.~Zihlmann, H.~Qin, and M.~Magno, ``Picosam2: Low-latency segmentation in-sensor for edge vision applications,'' \emph{IEEE Sensors Conference}, 2025.

\bibitem{Campanella_WearableSensor2024}
S.~Campanella, A.~Alnasef, L.~Falaschetti, A.~Belli, P.~Pierleoni, and L.~Palma, ``A novel embedded deep learning wearable sensor for fall detection,'' \emph{IEEE Sensors Journal}, 2024.

\bibitem{Vitolo_Low-PowerDetection2022}
P.~Vitolo, A.~De~Vita, L.~D. Benedetto, D.~Pau, and G.~D. Licciardo, ``Low-power detection and classification for in-sensor predictive maintenance based on vibration monitoring,'' \emph{IEEE Sensors Journal}, 2022.

\bibitem{moosmann2023ultraefficient}
J.~Moosmann, P.~Bonazzi, Y.~Li, S.~Bian, P.~Mayer, L.~Benini, and M.~Magno, ``Ultra-efficient on-device object detection on ai-integrated smart glasses,'' \emph{IEEE/CVF European Conference on Computer Vision (ECCV)}, 2023.

\bibitem{wang2020fann}
X.~Wang, M.~Magno, L.~Cavigelli, and L.~Benini, ``Fann-on-mcu: An open-source toolkit for energy-efficient neural network inference at the edge of the internet of things,'' \emph{IEEE Internet of Things Journal}, 2020.

\bibitem{giordano2022survey}
M.~Giordano, L.~Piccinelli, and M.~Magno, ``Survey and comparison of milliwatts micro controllers for tiny machine learning at the edge,'' \emph{IEEE International Conference on Artificial Intelligence Circuits and Systems (AICAS)}, 2022.

\bibitem{Kuhne_Resource-ConstrainedUAVs2025}
J.~Kühne, M.~Magno, and L.~Benini, ``Low latency visual inertial odometry with on-sensor accelerated optical flow for resource-constrained uavs,'' \emph{IEEE Sensors Journal}, 2025.

\bibitem{Leiva_IMU2025}
V.~Leiva, M.~Z.~U. Rahman, M.~A. Akbar, C.~Castro, M.~Huerta, and M.~T. Riaz, ``A real-time intelligent system based on machine-learning methods for improving communication in sign language,'' \emph{IEEE Access}, 2025.

\bibitem{Wei_GasConcentration2024}
G.~Wei, X.~Liu, A.~He, W.~Zhang, S.~Jiao, and B.~Wang, ``Design and implementation of a resnet-lstm-ghost architecture for gas concentration estimation of electronic noses,'' \emph{IEEE Sensors Journal}, 2024.

\bibitem{Cho_gas2024}
J.~Cho, Y.~J. Pyeon, Y.~M. Kwon, Y.~Kim, J.~Yeom, M.~W. Kim, C.~S. Park, I.-H. Kim, Y.~Lee, and J.~J. Kim, ``A mixture-gas edge-computing multisensor device with generative learning framework,'' \emph{IEEE Sensors Journal}, 2024.

\bibitem{Verma_AviEar2024}
R.~Verma and S.~Kumar, ``Aviear: An iot-based low-power solution for acoustic monitoring of avian species,'' \emph{IEEE Sensors Journal}, 2024.

\bibitem{zhou2024simulation}
Y.~Zhou, Y.~Luo, Z.~Yan, Y.~Jin, S.~Jiang, Z.~Wang, and B.~He, ``Simulation, design, and application of intelligent-edge-based soft magnetic tactile sensor with super-resolution,'' \emph{IEEE Sensors Journal}, 2024.

\bibitem{xie2024design}
Y.-L. Xie, X.-R. Lin, C.-Y. Lee, and C.-W. Lin, ``Design and implementation of an arm-based ai module for ectopic beat classification using custom and structural pruned lightweight cnn,'' \emph{IEEE Sensors Journal}, 2024.

\bibitem{MessinaQVAR_2025}
A.~Messina, A.~Lazzaro, R.~Carotenuto, and M.~Merenda, ``Preliminary analysis of the exploitation of qvar sensor for gesture recognition,'' \emph{International Conference on Smart and Sustainable Technologies (SpliTech)}, 2025.

\bibitem{imx500_sony_sensor}
\BIBentryALTinterwordspacing
``Sony imx500,'' 2023. [Online]. Available: \url{https://developer.sony.com/imx500/}
\BIBentrySTDinterwordspacing

\bibitem{eki2021sonyIMX500}
R.~Eki, S.~Yamada, H.~Ozawa, H.~Kai, K.~Okuike, H.~Gowtham, H.~Nakanishi, E.~Almog, Y.~Livne, G.~Yuval, E.~Zyss, and T.~Izawa, ``A 1/2.3inch 12.3mpixel with on-chip 4.97tops/w cnn processor back-illuminated stacked cmos image sensor,'' \emph{IEEE International Solid- State Circuits Conference (ISSCC)}, 2021.

\bibitem{bonazzi2023tinytracker}
P.~Bonazzi, T.~Rüegg, S.~Bian, Y.~Li, and M.~Magno, ``Tinytracker: Ultra-fast and ultra-low-power edge vision for in-sensor gaze estimation,'' \emph{IEEE Sensors Conference}, 2023.

\bibitem{zhou2020near}
F.~Zhou and Y.~Chai, ``Near-sensor and in-sensor computing,'' \emph{Nature Electronics}, 2020.

\bibitem{shu2023tinysam}
H.~Shu, W.~Li, Y.~Tang, Y.~Zhang, Y.~Chen, H.~Li, Y.~Wang, and X.~Chen, ``Tinysam: Pushing the envelope for efficient segment anything model,'' \emph{arXiv preprint arXiv:2312.13789}, 2023, also available via AAAI implementation.

\bibitem{zhou2023edgesam}
C.~Zhou, X.~Li, C.~C. Loy, and B.~Dai, ``Edgesam: Prompt-in-the-loop distillation for on-device deployment of sam,'' \emph{arXiv, 2312.06660}, 2023.

\bibitem{zhang2023mobilesamv2}
C.~Zhang, D.~Han, S.~Zheng, J.~Choi, T.-H. Kim, and C.~S. Hong, ``Mobilesamv2: Faster segment anything to everything,'' \emph{arXiv, 2312.09579}, 2023.

\bibitem{fu2024litesam}
J.~Fu, Y.~Yu, N.~Li, Y.~Zhang, Q.~Chen, J.~Xiong, J.~Yin, and Z.~Xiang, ``Lite-sam is actually what you need for segment everything,'' \emph{European Conference on Computer Vision (ECCV)}, 2024.

\bibitem{menghani2023efficient}
G.~Menghani, ``Efficient deep learning: A survey on making deep learning models smaller, faster, and better,'' \emph{ACM Computing Surveys}, 2023.

\bibitem{qin2024mobilenetv4}
D.~Qin, C.~Leichner, M.~Delakis, M.~Fornoni, S.~Luo, F.~Yang, W.~Wang, C.~Banbury, C.~Ye, B.~Akin \emph{et~al.}, ``Mobilenetv4: Universal models for the mobile ecosystem,'' \emph{European Conference on Computer Vision (ECCV)}, 2024.

\bibitem{wang2020eca}
Q.~Wang, B.~Wu, P.~Zhu, P.~Li, W.~Zuo, and Q.~Hu, ``Eca-net: Efficient channel attention for deep convolutional neural networks,'' \emph{IEEE/CVF Conference on Computer Vision and Pattern Recognition (CVPR)}, 2020.

\bibitem{zhang2023survey}
C.~Zhang, J.~Cho, F.~D. Puspitasari, S.~Zheng, C.~Li, Y.~Qiao, T.~Kang, X.~Shan, C.~Zhang, C.~Qin \emph{et~al.}, ``A survey on segment anything model (sam): Vision foundation model meets prompt engineering,'' \emph{arXiv, 2306.06211}, 2023.

\bibitem{dosovitskiy2020image}
A.~Dosovitskiy, ``An image is worth 16x16 words: Transformers for image recognition at scale,'' \emph{arXiv, 2010.11929}, 2020.

\bibitem{zhao2023fast}
X.~Zhao, W.~Ding, Y.~An, Y.~Du, T.~Yu, M.~Li, M.~Tang, and J.~Wang, ``Fast segment anything,'' \emph{arXiv, 2306.12156}, 2023.

\bibitem{zhang2023faster}
C.~Zhang, D.~Han, Y.~Qiao, J.~U. Kim, S.-H. Bae, S.~Lee, and C.~S. Hong, ``Faster segment anything: Towards lightweight sam for mobile applications,'' \emph{arXiv, 2306.14289}, 2023.

\bibitem{liang2022expediting}
W.~Liang, Y.~Yuan, H.~Ding, X.~Luo, W.~Lin, D.~Jia, Z.~Zhang, C.~Zhang, and H.~Hu, ``Expediting large-scale vision transformer for dense prediction without fine-tuning,'' \emph{Advances in Neural Information Processing Systems}, 2022.

\bibitem{chen20230}
Z.~Chen, G.~Fang, X.~Ma, and X.~Wang, ``0.1\% data makes segment anything slim,'' \emph{arXiv, 2312.05284}, 2023.

\bibitem{lv2024ptq4sam}
C.~Lv, H.~Chen, J.~Guo, Y.~Ding, and X.~Liu, ``Ptq4sam: Post-training quantization for segment anything,'' \emph{IEEE/CVF Conference on Computer Vision and Pattern Recognition (CVPR)}, 2024.

\bibitem{liu2023pqsam}
X.~Liu, X.~Ding, L.~Yu, Y.~Xi, W.~Li, Z.~Tu, J.~Hu, H.~Chen, B.~Yin, and Z.~Xiong, ``Pq-sam: Post-training quantization for segment anything model,'' \emph{European Conference on Computer Vision (ECCV)}, 2024.

\bibitem{banner2019post}
R.~Banner, I.~Hubara, E.~Hoffer, and D.~Soudry, ``Post training 4-bit quantization of convolutional networks for rapid-deployment,'' \emph{Advances in Neural Information Processing Systems (NeurIPS)}, 2019.

\bibitem{nagel2020up}
M.~Nagel, M.~van Baalen, T.~Blankevoort, and M.~Welling, ``Up or down? adaptive rounding for post-training quantization,'' \emph{European Conference on Computer Vision (ECCV)}, 2020.

\bibitem{ryali2023hiera}
C.~Ryali, Y.-T. Hu, D.~Bolya, C.~Wei, H.~Fan, P.-Y. Huang, V.~Aggarwal, A.~Chowdhury, O.~Poursaeed, J.~Hoffman \emph{et~al.}, ``Hiera: A hierarchical vision transformer without the bells-and-whistles,'' \emph{International Conference on Machine Learning (ICML)}, 2023.

\bibitem{he2022masked}
K.~He, X.~Chen, S.~Xie, Y.~Li, P.~Doll{\'a}r, and R.~Girshick, ``Masked autoencoders are scalable vision learners,'' \emph{IEEE/CVF Conference on Computer Vision and Pattern Recognition (CVPR)}, 2022.

\bibitem{xu2023auto}
Y.~Xu, F.~Yu, and B.~Zhou, ``Autoppn: Learning to design promptable pyramid networks for efficient segmentation,'' \emph{IEEE/CVF International Conference on Computer Vision (ICCV)}, 2023.

\bibitem{wang2023litevit}
H.~Wang, K.~Wang, Y.~Xu, C.~Xu, and C.~Shen, ``Litevit: Towards efficient vision transformers with enhanced token mixing,'' \emph{arXiv, 2303.13429}, 2023.

\bibitem{bonazzi2026tinyglass}
P.~Bonazzi, R.~Sutter, L.~Capogrosso, articleha Buob, and M.~Magno, ``Tinyglass: Real-time self-supervised in-sensor anomaly detection,'' \emph{arXiv, 2603.16451}, 2026.

\bibitem{Tong2024}
Q.~Tong, J.~Wang, W.~Yang, S.~Wu, W.~Zhang, C.~Sun, and K.~Xu, ``{Edge AI-enabled chicken health detection based on enhanced FCOS-Lite and knowledge distillation},'' \emph{Computers and Electronics in Agriculture}, 2024.

\bibitem{Cui2024}
T.~Cui, Z.~Zhang, C.~Sun, S.~Wang, H.~Li, and W.~Zhang, ``{Pedestrian Warning: Intelligent Vision Sensor vs. Edge AI with LTE C-V2X in a Smart City},'' \emph{IEEE 99th Vehicular Technology Conference (VTC2024-Spring)}, 2024.

\bibitem{ronneberger2015unet}
O.~Ronneberger, P.~Fischer, and T.~Brox, ``{U-Net}: Convolutional networks for biomedical image segmentation,'' 2015.

\bibitem{chollet2017xception}
F.~Chollet, ``Xception: Deep learning with depthwise separable convolutions,'' \emph{IEEE Conference on Computer Vision and Pattern Recognition (CVPR)}, 2017.

\bibitem{hinton2015distilling}
G.~Hinton, O.~Vinyals, and J.~Dean, ``Distilling the knowledge in a neural network,'' \emph{arXiv, 1503.02531}, 2015.

\bibitem{milletari2016v}
F.~Milletari, N.~Navab, and S.-A. Ahmadi, ``V-net: Fully convolutional neural networks for volumetric medical image segmentation,'' \emph{International Conference on 3D Vision (3DV)}, 2016.

\bibitem{milletari2016vnet}
F.~Milletari, N.~Navab, and S.~Ahmadi, ``V‑net: Fully convolutional neural networks for volumetric medical image segmentation,'' \emph{International Conference on 3D Vision (3DV)}, 2016.

\bibitem{Sanh2019DistilBERTAD}
V.~Sanh, L.~Debut, J.~Chaumond, and T.~Wolf, ``Distilbert, a distilled version of bert: smaller, faster, cheaper and lighter,'' \emph{arXiv, 1910.01108}, 2019.

\bibitem{loshchilov2019decoupledweightdecayregularization}
I.~Loshchilov and F.~Hutter, ``Decoupled weight decay regularization,'' \emph{arXiv}, 2019.

\bibitem{lin2014microsoft}
T.-Y. Lin, M.~Maire, S.~Belongie, L.~Bourdev, R.~Girshick, J.~Hays, P.~Perona, D.~Ramanan, C.~L. Zitnick, and P.~Doll{\'a}r, ``Microsoft coco: Common objects in context,'' \emph{arXiv, 1405.0312}, 2014.

\bibitem{habi2021hptq}
H.~V. Habi, R.~Peretz, E.~Cohen, L.~Dikstein, O.~Dror, I.~Diamant, R.~H. Jennings, and A.~Netzer, ``Hptq: Hardware-friendly post training quantization,'' \emph{arXiv, 2109.09113}, 2021.

\bibitem{gordon2024eptq}
O.~Gordon, E.~Cohen, H.~V. Habi, and A.~Netzer, ``Eptq: Enhanced post-training quantization via hessian-guided network-wise optimization,'' \emph{European Conference on Computer Vision (ECCV) Workshops}, 2024.

\bibitem{dikstein2025dgh}
L.~Dikstein, A.~Lapid, A.~Netzer, and H.~V. Habi, ``Data generation for hardware-friendly post-training quantization,'' \emph{IEEE/CVF Winter Conference on Applications of Computer Vision (WACV)}, 2025.

\bibitem{gholami2021surveyquantizationmethodsefficient}
\BIBentryALTinterwordspacing
A.~Gholami, S.~Kim, Z.~Dong, Z.~Yao, M.~W. Mahoney, and K.~Keutzer, ``A survey of quantization methods for efficient neural network inference,'' 2021. [Online]. Available: \url{https://arxiv.org/abs/2103.13630}
\BIBentrySTDinterwordspacing

\end{thebibliography}
